\documentclass[a4paper, times, 10pt, twocolumn, twoside]{article}

\usepackage{ISARC}

\usepackage{lscape}
\usepackage{hologo}

\usepackage{algorithm}
\usepackage{algorithmic}
\usepackage{accents} 
\usepackage{soul,color}

\begin{document}

\linespread{0.5}

\title{Stag hunt game-based approach for cooperative UAVs}

\author{L.V. Nguyen$^{1}$, I.Torres Herrera$^{1}$, T.H. Le$^{1}$, M.D. Phung$^{1,2}$, R.P. Aguilera$^{1}$ and Q.P. Ha$^1$}

\affiliation{
$^1$School of Electrical and Data Engineering, University of Technology Sydney
(UTS), Australia\\
$^2$VNU University of Engineering and Technology (VNU-UET), Vietnam National
University, Hanoi (VNU), Vietnam
}

\email{
\href{mailto:e.author1@aa.bb.edu}{\{vanlanh.nguyen; ignacio.torresherrera; hoang.t.le; manhduong.phung;
		ricardo.aguilera; quangha\}@uts.edu.au}
}

\maketitle 
\thispagestyle{fancy} 
\pagestyle{fancy}

\begin{abstract}
Unmanned aerial vehicles (UAVs) are being employed in many areas such as photography, emergency, entertainment, defence, agriculture, forestry, mining and construction. Over the last decade, UAV technology has found applications in numerous construction project phases, ranging from site mapping, progress monitoring, building inspection, damage assessments, and material delivery. While extensive studies have been conducted on the advantages of UAVs for various construction-related processes, studies on UAV collaboration to improve the task capacity and efficiency are still scarce. This paper proposes a new cooperative path planning algorithm for multiple UAVs based on the stag hunt game and particle swarm optimization (PSO). First, a cost function for each UAV is defined, incorporating multiple objectives and constraints. The UAV game framework is then developed to formulate the multi-UAV path planning into the problem of finding payoff-dominant equilibrium. Next, a PSO-based algorithm is proposed to obtain optimal paths for the UAVs. Simulation results for a large construction site inspected by three UAVs indicate the effectiveness of the proposed algorithm in generating feasible and efficient flight paths for UAV formation during the inspection task.
\end{abstract}


\section{Introduction}
\label{sec:Introduction}

The immense development of unmanned aerial vehicles (UAVs) technologies has been drawing significant attention in civilian sectors. In the construction domain, researchers and enterprises tend to seek safer and more efficient solutions for carrying out construction-related tasks. As modern technologies have reduced the cost of UAVs while increasing their dependability, operating time, and maneuverability, smart drone-powered solutions have been introduced as a platform to assist construction activities. They are well-established in numerous construction project stages such as construction site monitoring and 3D mapping \cite{bulgakov2020generation}, building and damage inspection \cite{bolourian2017high}, and package delivery logistics \cite{grzybowski2020low}, demonstrating prospects for wide usage of drones.

Due to the increased quantity and complexity of construction jobs, such as large-site 3D mapping or multiple-package delivery, a single drone with restricted size and capability may sometimes not fulfill the requirements. Consequently, multi-UAVs can be deployed to collaborate as a team for the applications mentioned above in order to optimize processing time and operating possibilities \cite{real2021experimental}.

\begin{figure}[h!]
	\begin{center}
		\includegraphics[width = \linewidth]{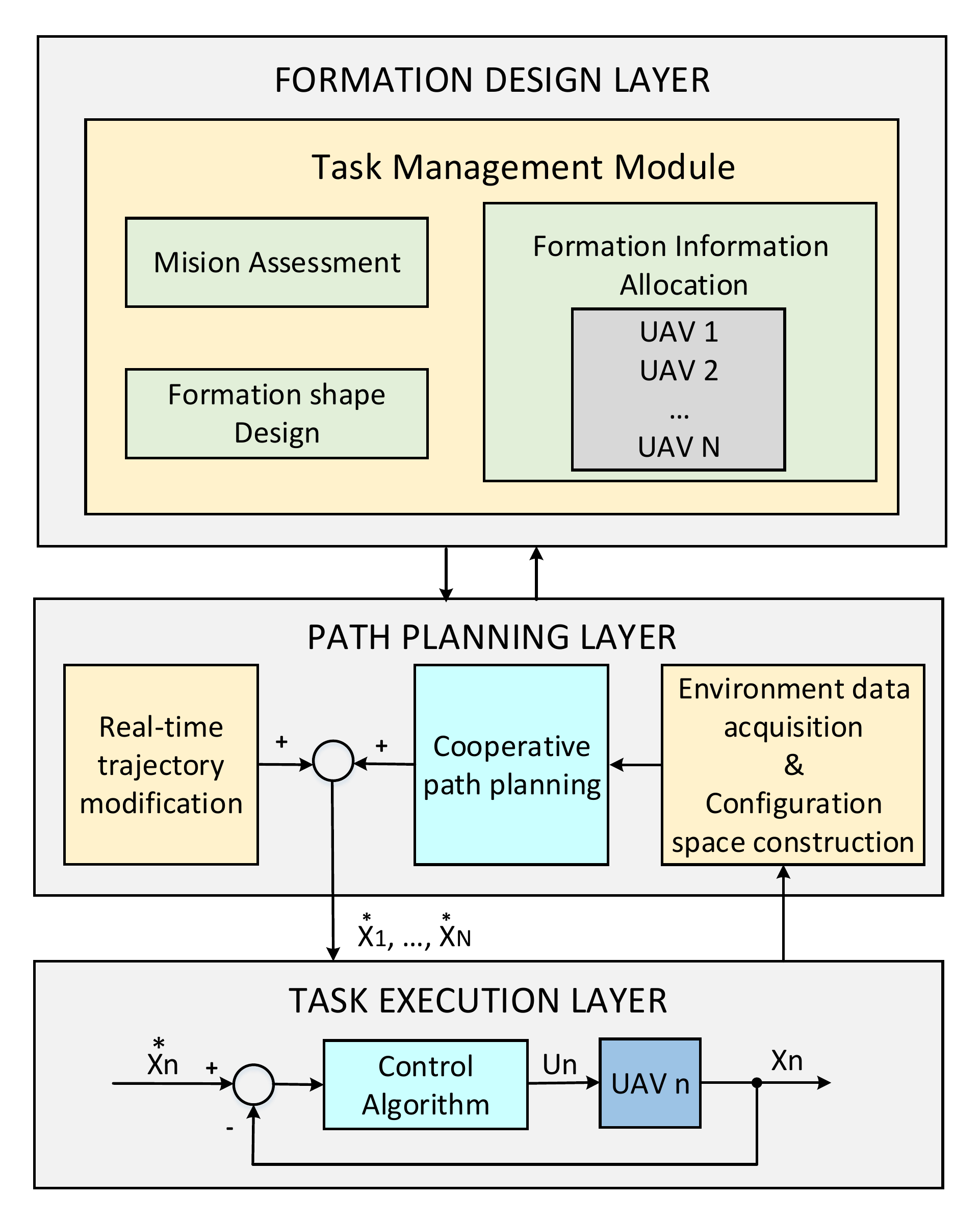}
		\caption{Hierarchical UAV control structure.}
		\label{fig01}
	\end{center}
\end{figure}

A hierarchical structure formation system includes three layers: task management, path planning, and task execution, as shown in Figure \ref{fig01}. The task management layer holds and keeps track of the mission's objectives. Based on these objectives, this layer allocates resources and tasks to UAVs and acts as a decision-maker. From mission requirements, the path planning layer generates feasible trajectories for the formation. This layer aims to plan optimal paths for the UAVs moving in a known environment from the start to their target locations. This layer comprises three blocks: real-time trajectory modification, data acquisition, and cooperative path planning, wherein the collaborative path planning block is the primary function of the system and determines the overall optimized path for each quadcopter. Nonetheless, due to uncertainties that may occur along the route in practical applications, the real-time trajectory modification block is combined with the system. The formation can deal with emergencies such as a suddenly appearing obstacle. Generated paths will then be passed down to the task execution layer. This layer directly is connected with the propulsion system of the quadcopter and generates the control laws. To enhance system performance, real-time data, i.e., UAVs' position and velocity, are fed back to the path planning layer to adjust the path, providing a closed control loop. 

This paper focuses on cooperative path planning, where the path-planner produces trajectories to fulfill the mission objectives. The mission objectives include formation shape maintenance, minimum path lengths, and threat avoidance. The development is then illustrated in the inspection of a construction site.

\section{Related works and Rationale}
Path planning for multiple UAVs has been becoming an active research topic recently. The objectives and approaches may be different depending on the application domain. 

The artificial potential field is one of the most widely used techniques for UAV cooperative path planning \cite{mac2018development}. It considers the operating space as a potential field, with attractive fields surrounding the target and repulsive fields around obstacles. For cooperation, new fields are added, including the internal attractive potential fields to retain the formation configuration and the internal repulsive fields to prevent the UAVs from colliding with each other. The paths are then generated under the total force acting on the UAVs at each position. This technique can produce smooth and continuous paths. It, however, faces local minima problems when the total force is reaching zero.

In another direction, optimal control methods have been used for cooperative path planning by considering a numerical optimization problem subject to multiple constraints \cite{kendoul2012survey}. It first finds paths for single vehicles and then attains their cooperation by constraints. To generate paths, optimal control signals for individual quadcopters are computed using the mixed-integer linear programming (MILP). Although the MILP can solve the optimization problem with different constraints, it involves high computational complexity.

Recently, evolutionary algorithms (EA) have been used to solve multi-UAV cooperative path planning with the capability to find optimal solutions in complex scenarios \cite{kala2012multi}. They often include two stages, one for individual UAV path planning and the other for path cooperation. Initially, each vehicle is associated with the EA process to generate a feasible path. Those paths are then adjusted via a global cost function to achieve the required cooperation. This approach can generate smooth cooperative paths as it does not discretize the workspace. It however may converge to sub-optimal solutions if cooperative constraints and maneuver properties of UAVs are not properly addressed.

On the other hand, as a theoretical framework for strategic resolving of interactions among competing players, game theory has gained its applications in a variety of fields such as construction bidding \cite{ahmed2016construction} and environmental management in the mining industry \cite{collins2020game}. In the game, the purpose is to maximize the profit, which depends on not only actions of a player but also other players. Therefore, the best strategy relies on what the player expects others to do. Most games in the literature can be classified as cooperative and non-cooperative \cite{moura2018game}. In cooperative games (CGs), several players share a common objective to better achieving it than those working alone. However, a major issue with CGs remains trade-off between the stability of the player groups and the system efficiency \cite{han2012game}. In contrast, each player in noncooperative games (NCGs) has their own properties such as the payoff function, procedural details of the game, intention, and possible strategies which make him more inclusive than in CGs. Among NCGs, Nash Game is widely applied to the situation that all players have to simultaneously make their decisions in symmetric competitions, such as in exploring the public-private partnership investment incentives \cite{ho2017analysis} or seeking strategies for clusters in a distributed system \cite{zeng2019generalized}. In these games, each player only has partial information about the choices of others.

In general, Nash games revolve around the prisoner's dilemma (PD), where the stable cooperation is inhibited by its most restrictive conditions: defecting is a dominating option as it always offers a greater payout regardless of how the other player while collaboration is risky \cite{macy2002learning}. Meanwhile, numerous types of dilemmas are available, the most famous of which is the stag hunt game \cite{skyrms2004stag}, therein players desire to coordinate, i.e., the preferred choice is to always do as the rival acts. While in such games, mutual defection and mutual cooperation are evolutionary stable strategies, only the mutual cooperation results in a pay-off dominant equilibrium \cite{gintis2009game}.

From the UAVs' cooperation perspective, where profit from cooperation is much more than profit from individual effort, the stag hunt game can better resolve the conflict between individual UAVs. This is especially the case when examining the influence of group selection, in which social interactions aim to maximize the group's performance.

This paper proposes a stag hunt game based algorithm for UAV cooperative path planning. A cost function is first defined including all requirements on formation, path feasibility, safety and optimality. Unlike existing EA approaches, our method considers cooperative constraints in every individual to maximize the overall profit. The path planning problem is therefore modelled as a game where UAVs are the players. The strategy for each UAV is then formulated and enhanced PSO is introduced to obtain the payoff-dominant equilibrium. As a result, optimal paths can be achieved with the formation being maintained.

This paper is structured as follows. Section 3 formulates the cooperative path planning problem. Section 4 presents the proposed stag hunt based game and enhanced PSO. Numerical simulation results are provided in Section 5. Finally, a conclusion is given in Section 6.

\section{Problem formulation}
\subsection{Multi-vehicle path planning}
\begin{figure}[h!]
	\begin{center}
		\includegraphics[width= 0.95\linewidth]{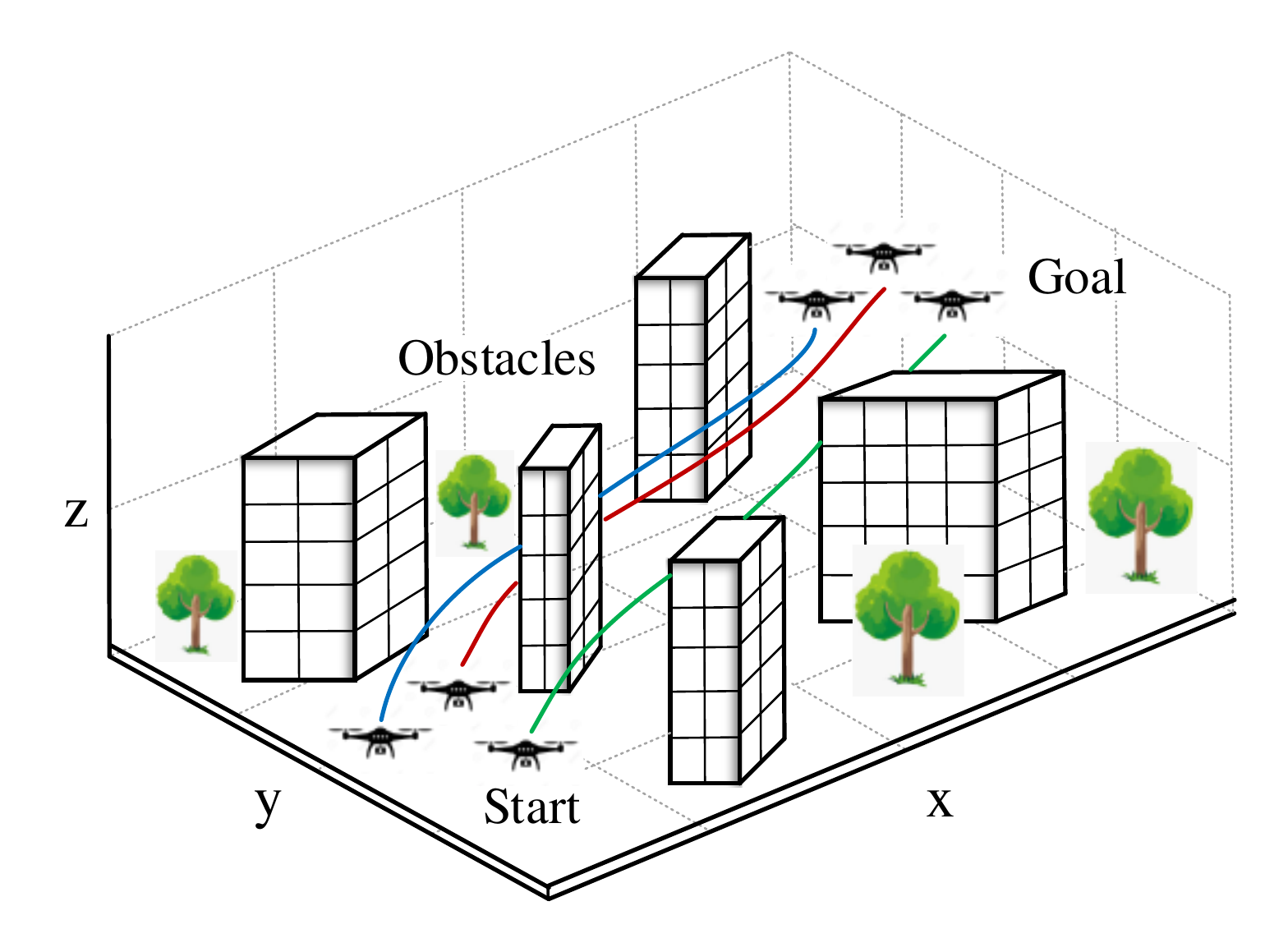}
		\caption{Definition of the path planning problem.}
		\label{fig02}
	\end{center}
\end{figure}

Consider a team of $N$ drones operating in a given flying area, including numerous obstacles, as shown in Figure \ref{fig02}. The position of the UAV group is determined in an earth frame, $xyz$, by $P \!=\! [{P_1}^T, {P_2}^T,...,{P_N}^T]^T$, where $N$ is the total number of drone members and $P_n \!=\! (x_n, y_n, z_n)^T$ is the location of the $n$-th vehicle. 

The problem of path planning is to establish a feasible route connecting the start and target positions in a collision-free environment while fulfilling a number of constraints. The problem can be expressed as an optimization process that is subjected to several costs. For a single UAV planning problem, it can be formulated as
\begin{equation}\label{eq01}
P(0)  \overset{X(k)}{\underset{s.t. ~ J_{s}(X(k))}{\xrightarrow{\hspace*{1cm}}}} P(end),
\end{equation}
where $P(0) $ and $P(end)$ are corresponding to the start and the target poses, $k$ stands for the waypoint instant, and $X(k)$ denotes the path of the UAV, including a total of $K$ waypoints subjected to the single-UAV cost $J_{s}(X(k))$.

In single-vehicle path planning, to attain the most effective path, the cost  $J_{s}(X(k))$ should be optimized, fulfilling constraints on path length, threat avoidance, and turning angle limit. It is defined as
\begin{equation}
\begin{split}\label{eq02}
 J_s(X(k)) & =  \omega_1 \sum_{k=1}^{K-1} L(k) +  \omega_2 \sum_{k=1}^{K-1} \sum_{\tau =1}^{\mathcal{T}} D_\tau(k) + \omega_3 \sum_{k=1}^{K} H(k) \\
& \quad +  \omega_4\sum_{k=1}^{K-2} \theta(k) + \omega_5\sum_{k=1}^{K-1} \left| \varphi(k) - 	\varphi(k+1) \right |,
\end{split}
\end{equation}
where $L(k)$ is the path length, $D(k)$ is the safety cost concerning $\mathcal{T}$ threats, $H(k)$ stands for the altitude payoff, $\theta_n(k)$ and $\varphi_n(k)$ correspond to the turning angle and climbing angle, $\omega_i$, for $i = \{1,2,...,5\}$, are the weight coefficients. More details about the single cost function are presented in our previous work \cite{phung2021safety}.

Extending the problem into the multi-vehicle formation path planning, it is written as
\begin{equation}\label{eq03}
P_n(0)  \overset{X_n(k)}{\underset{s.t. ~ J(X_n(k), X_n^-(k))}{\xrightarrow{\hspace*{1cm}}}} P_n(end), ~ n = 1,2,...,N,
\end{equation}
where $X_n(k)$ and $X_n^-(k)$ are corresponding to the path of UAV$_n$ and a set of its neighbours' paths. The multi-vehicle cost function, $J(X_n(k), X_n^-(k))$, consists of a single cost and a formation cost, computed as
\begin{equation}\label{eq04}
J(X_n(k), X_n^-) = J_{s}(X_n) + \beta J_{f}(X_n, X_n^-),
\end{equation}
where $J_{s}(X_n)$ is computed as \eqref{eq02}, $J_{f}(X_n, X_n^-)$ is the formation cost, and $\beta$ is a weighting factor. The cost function for the formation constraint is determined as follows.

\subsection{Formation cost function}
We apply here the graph-theoretic approach to represent the structure of the formation and interaction among UAVs. A graph is defined by $\mathcal{G} \!=\! (\mathcal{V}, \mathcal{E})$, in which $\mathcal{V} \!=\! \{v_1, v_2, …,v_N\}$ and $\epsilon \!=\! {(v_n,v_{n'}) \in \mathcal{E}}$ represent the vehicles in the group and their interconnections, respectively. To form a formation, there must exists an interconnection in $\mathcal{E}$ between any two vertices $(v_n,v_{n'}) \!\in\! \mathcal{V}$. Consider our graph is egde-weighted, i.e., each interconnection in the graph is weighted by $\mu_{nn'}$. The graph incidence matrix $\mathcal{D}$ has the dimension of $N\times M$, where its $uv$-th entry is equal to $1$ or $-1$ if the UAV$_n$ is respectively the head or tail  of the $v$-th edge, and $0$ otherwise.

The configured formation shape is defined by a reference position set, $P_r = [{P_{1_r}}^T, {P_{2_r}}^T,...,{P_{S_r}}^T]^T$,  where an element  $P_{n_r} = (x_{n_r}, y_{n_r}, z_{n_r})^T$ is given as the reference position of UAV$_n$. The relative reference between two neighbors, $n$ and $n'$, is then obtained as $P_{nn'_r} = P_{n_r} - P_{n'_r}$. Accordingly, the formation error between UAV$_n$ and  UAV$_n'$ is calculated as $P_n - P_{n'} - P_{nn'_r}$. Denote $ \mathcal{\hat{D}} \!=\! \mathcal{D} \otimes I_3$, where operator $\otimes$ is the Kronecker product. The formation error for UAV$_n$ can be expressed as
\begin{equation}\label{eq05}
\begin{split}
&E_n=\sum_{n' \in \mathcal{E}}  \mu_{nn'} \|P_n - P_{n'} -  P_{nn'_r} \|^2 \\
&  = (P - P_r)^T \mathcal{\hat{D}} \hat W_n \mathcal{\hat{D}}^T (P - P_r) = ||P - P_r||^2_{ \mathcal{\hat{D}} \hat W_n \mathcal{\hat{D}}^T},
\end{split}
\end{equation}
where $ \hat W_n = W_n \otimes I_3  $ and  $ W_n = \text{diag}[\mu_{nn'}]$ is a diagonal matrix of dimension $M \times M$ for the weights $\mu_{nn'}.$

Let $\bar d_n(k)$ be the Euclidean distance from UAV$_n$ to its nearest neighbor at the waypoint $k$, $r_n$ be the radius of UAV$_n$, and $d_s$ be the safe distance. To avoid collision between vehicles, the distance between a UAV and its nearest neighbor needs to be smaller than the sum of a safe distance, $d_s$, and twice the UAV radius, $r_n$. Therefore, we incorporate this collision avoidance condition by reformulating the formation error as 
\begin{equation}\label{eq06}
\begin{split}
& E_n(k)  = 	\begin{cases}
\begin{split}
||P(k) -  P_r || ^2_{ \mathcal{\hat{D}} \hat W_n \mathcal{\hat{D}}^T}, ~   & \text{if} \;  \bar d_n(k) > d_s + 2r_n\\	
\infty,\qquad \qquad ~   & \text{if} \;  \bar d_n(k) \leq d_s  + 2r_n.
\end{split}
\end{cases}
\end{split}
\end{equation}
The formation cost function is then defined as below:
\begin{equation}\label{eq07}
J_{f}(X_n, X_n^-) = \sum_{k=1}^{K}E_n(k).
\end{equation}

\section{Stag hunt game-based algorithm for UAV cooperative path planning}
Given the cost function $J(X_n, X_n^-)$ defined for each UAV, the cooperative path planning becomes finding paths $X_n, n = 1,2,...,N$ to simultaneously minimize $J(X_n, X_n^-)$. Since this cost depends on not only path $X_n$ generated for UAV$_n$ but also its rivals' paths $X_n^-$, finding optimal solutions is a challenging problem.

\subsection{The game of stag hunt}
To resolve conflicts and interactions among rational decision-makers  \cite{ji2021game}, players can pursue their individual objectives by considering possible goals, behaviors, and countermeasures of other decision-makers to achieve a win-win situation.

\begin{figure}[h!]
	\begin{center}
		\includegraphics[width= 0.9\linewidth]{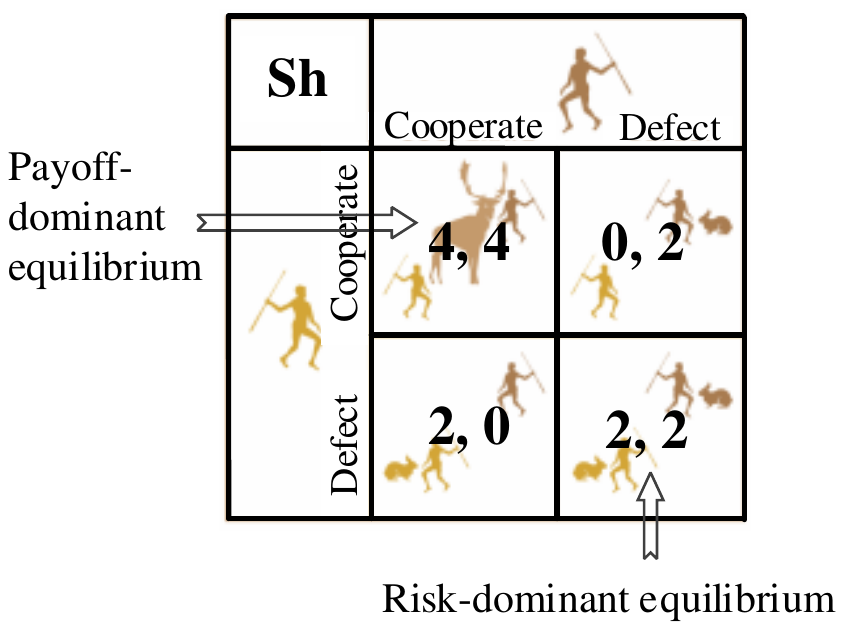}
		\caption{Payoff matrix of stag hunt game}
		\label{fig03}
	\end{center}
\end{figure}

The stag hunt game, originated by J.J. Rousseau \cite{rousseau1754discourse}, illustrates a conflict between safety and social cooperation. In the game, two hunters independently decide whether to hunt a stag or a hare without knowing the other's decision. One hunter can catch a hare individually with a high guarantee of success. Meanwhile, the value of a shared stag is far greater than that of a hare, but cooperation between hunters is required to hunt a stag successfully. Therefore, it would be much better for each hunter to choose a more ambitious and far more rewarding goal instead of deciding on a total autonomy and minimal risk strategy. The payoff matrix in Figure \ref{fig03} illustrates a generic stag hunt with two players.

In a game, "Nash equilibrium" is reached when no player can obtain more profits if others do not change their strategies. Formally, a stag hunt is a game with two pure Nash equilibria: risk dominant and payoff dominant. A Nash equilibrium is called "risk dominant" if it has the largest basin of attraction, implying that it is less risky. This means that the more ambiguity players have about the other player's intentions, the more likely they are to pick the plan that best suits them. Meanwhile, the payoff dominant equilibrium is defined as being Pareto superior to all other Nash equilibria in the game.

Pareto optimality is a fundamental concept representing efficiency in a multi-objective optimization problem consisting of several conflicting objectives. A set of alternatives is considered a Pareto optimal solution if no reallocation can further improve any one of the objectives without degrading at least one other. In the stag hunt game, when confronted with a decision among multiple equilibria, all players would vote on the payoff dominant equilibrium since it provides each member with at least as much profit as the other Nash equilibria. 

From the mathematical point of view, a formal presentation of the payoff-dominant equilibrium is as follows. Consider a stag hunt game described as $G = (N, S, U),$ where $N$ is a set of players, $S \!=\! (S_1, S_2,..., S_N)$ denotes strategy sets, where $S_n \!=\! (x_{n_1}, x_{n_2},..., x_{n_\Sigma}) , ~ n \!=\! 1,..., N$, represents all $\Sigma$ strategies made by the $n$-th player, and $U = (U(x_1), U(x_2),...,U(x_N))$
 stands for a set of players' utility. The allocation $\accentset{\ast}{X} = \{\accentset{\ast}{x_1},\accentset{\ast}{x_2},...,\accentset{\ast}{x_N} \}$, where $\accentset{\ast}{x_n} \in S_n$, is defined as Pareto optimal if it dominates all other reallocation $X = \{x_1, x_2,...,x_N\}$, i.e., both of the following requirements are met:
\begin{subequations}\label{eq09}
	\begin{equation}
	\forall n\in \{1,...,N\}, U(\accentset{\ast}{x_n}) \geq  U(x_n),
	\end{equation}
	\begin{equation}
	\nexists n' \in \{1,...,N\}: U(\accentset{\ast}{x}_{n'}) <  U(x_{n'}).
	\end{equation}
\end{subequations} 

\subsection{The game of UAV cooperation}
Each vehicle in the cooperative path planning problem has been assigned a cost function $J(X_n, X_n^-)$ defined in \eqref{eq04}. These cost functions interact among vehicles. Therefore, it becomes challenging to solve multiple optimal problems simultaneously. As can be seen, the stag hunt game concept aligns well with the cooperative path planning problem and is promising for its solution. As such, this paper proposes a stag hunt game-based approach for cooperative UAVs. The two-step procedure to implement the proposed game-based scheme is as follows.

In the first step, the cooperation of multiple UAVs is formulated as a game to model interactions among the drones, including three key elements: players, strategies, and utility. Each vehicle in the formation is considered as a player, also called a decision-maker. During the game, all UAV players have to simultaneously provide a route, $X_n$, defined as the player's strategy, without knowledge of the other player's decision. Each player will get his own utility, corresponding to the multi-vehicle cost  $J(X_n, X_n^-)$, which is a function of strategies made by himself $X_n$ and his rivals, $X_n^-$.

Indeed, one UAV can reach its target position alone by solving its single UAV path planning problem to obtain the minimum single cost, $J_s(X_n)$. In a cooperative task, however, this individual optimal solution could result in a significant formation error, $E_n(k)$, leading to a high formation cost  $J_f(X_n, X_n^-)$. To successfully perform a cooperative UAV mission with a higher profit, it would be better for each player to choose the more ambitious goal of achieving a far greater reward by providing a formation preserving path in expectation of the other vehicle's cooperation. Therefore, the multi-vehicle cost, $J(X_n, X_n^-)$, combining both single cost $J_s(X_n)$ and formation cost  $J_f(X_n, X_n^-)$, should be optimized for all players simultaneously. Accordingly, the game aims to find a payoff-dominant equilibrium.

In the second step, an enhanced PSO-based algorithm is introduced to solve the Pareto optimality, resulting in a payoff-dominant equilibrium as a desired outcome. This step will be presented in the following section.

\subsection{Enhanced PSO-based approach for finding payoff-dominant equilibrium}
Particle swarm optimization (PSO) is a stochastic optimization algorithm for optimizing a problem by iteratively improving a candidate solution concerning a particular quality measure. It solves a problem by generating a population of possible solutions, known as particles, and relocating them in the search space using a few simple formulae based on the particle's position and velocity. Each particle's movement is guided by its local best-known position and the global best-known pose in the search space, updated when other particles discover better places. This is anticipated to direct the swarm toward the best options.

Formally, consider a $d$-dimension search space and a swarm consisting of $N_{pop}$ particles, each particle $i$ has a position $X_i \in \mathbb{R}^d$ and a velocity $V_i \in \mathbb{R}^d$. Let $Q_i$ be the best known position of particle $i$ and $Q_g$ be the best known position of the entire swarm. The dynamic algorithm of the swarm is defined as below:
\begin{equation}\label{eq010}
V_i(t+1)  = c_0 V_i(t) + c_1r_1[Q_i(t) - X_i(t)] +c_2r_2[Q_i(t) - X_i(t)],
\end{equation}
\begin{equation}\label{eq011}
X_i(t+1) = X_i(t) + V_i(t+1),  \qquad  \qquad  \qquad  \qquad  \qquad  \;
\end{equation}
where $c_0$ is the inertia weight, $c_1$ and $c_2$ are corresponding to self confidence and swarm confidence parameters, and $r_1$ and $r_2$ are random values uniformly distributed in the interval $[0,1]$.

In the UAV path planning problem, the position of a particle is encoded by the flight path $X_n$. Accordingly, the entire swarm consists of $N_{pop}$ path particles, which are updated for the optimal solution. To speed up the search process, we employ in this study a variant of PSO named spherical vector-based particle swarm optimization (SPSO) developed in our previous work \cite{phung2021safety}. In the SPSO, waypoints of a flight path are represented in the spherical coordinate system to exploit the corresponding magnitude, elevation, and azimuth components of the variables with speed, turning angle, and climbing slope of the UAV.  
	
To further develop the SPSO for cooperative path planning involving multiple UAVs, we introduce a game-based SPSO to find the payoff-dominant equilibrium. The pseudo-code for the optimization process is described in Algorithm \ref{al1}. The detail of the algorithm is as follows.

\begin{algorithm}
	\caption{Game-based PSO implementation}
	\begin{algorithmic} 
		\STATE 1. Initialize PSO parameters: $c_{0}, c_{1}, c_{2}, maxIt, N_{pop}$;
		\STATE 2. Set $It = 0$, generate random player's strategies; 
		\STATE 3. Obtain the initial optimal strategies $\accentset{\ast}{X}_n$, $\accentset{\ast}{X}_{n}^-$;
		\FOR{$It= 1:maxIt$}
		\STATE 4. Calculate	$J(X_{n}(It), {X}_{n}^-(It))$, for $n = 1, 2, ..., N$;
		\IF{$J(X_{n}(It), {X}_{n}^-(It)) \leq J({\accentset{\ast}{X}_{n},\accentset{\ast}{X}_{n}^-)}$, $\forall n = 1, 2, ..., N$}
		\STATE 5. Update  $\accentset{\ast}{X}_{n} = X_{n}(It); \accentset{\ast}{X}_{n}^- = {X}_{n}^-(It)$;
		\ENDIF
		\STATE 6. Record  $\accentset{\ast}{X}_{n}$, $\accentset{\ast}{X}_{n}^-$;	
		\STATE 7. Update $X_{n}$, ${X}_{n}^-$;
		\ENDFOR	
		\STATE 8. Obtain $\accentset{\ast}{X}_n$,  $\forall n = 1, 2, ..., N$.
	\end{algorithmic}
	\label{al1}
\end{algorithm}

(i) \textit{Initialization:}

Initially, parameters of the PSO including $c_0, c_1, c_2$, number of iterations $maxIt$, and number of particles $N_{pop}$ are first initialized. At this stage, corresponding to $It = 0$, random strategies of the players are also generated and assigned as initial optimal strategies $\accentset{\ast}{X}_n$, $\accentset{\ast}{X}_{n}^-$.

(ii) \textit{Evaluation:}

At each iteration, from $It \!=\! 1$ to $maxIt$, cost values representing the players' profit, $J_i( X_{n}(It), {X}_{n}^-(It))$, are computed as \eqref{eq04}, where $i$ denotes a particle in the $n$-th swarm.

(iii) \textit{Optimal strategy:}

The best strategies of all players associated to the particle $i$ at the iteration $It$ are updated if there is more benefit for at least one player without decreasing the other players' profit, i.e., the condition \eqref{eq09} is met. Based on them, the strategy of each player is adjusted for the subsequent iteration according to equations (\ref{eq010}) and (\ref{eq011}) of the PSO. 

(iv) \textit{Termination:}

The algorithm is terminated when exceeding the maximum number of iterations, $maxIt$. Finally, the payoff-dominant equilibrium, or the best allocation, $ \accentset{\ast}{X} \!=\! \{\accentset{\ast}{X}_1, \accentset{\ast}{X}_2,..., \accentset{\ast}{X}_N \}$, is obtained.

\section{Simulation results}
This section presents simulation results of the proposed path planning algorithm for a group of three drones. The aim is to generate paths for the UAVs flying in an equilateral triangle formation. However, it should be noted that there are no specific constraints on the configuration of formation shapes. The incidence matrix $\mathcal{D} $ is defined as
\begin{equation}\label{eq026}
\mathcal{D} = \begin{bmatrix}
1 & 1 & 0  \\
-1 & 0 & 1  \\
0 & -1 & -1  \\
\end{bmatrix}.
\end{equation}
The interconnection weights are set as $W_1 \!=\! [1 ~ 1 ~ 0]$, $W_2 \!=\! [1 ~ 0 ~ 1]$, and $W_3 \!=\! [0 ~ 1 ~ 1]$. Noting that the weights for interconnections are set equal among players since all players play a similar role in the team. In the total cost function, weighting coefficients $[\beta, ~\omega_{1},~ \omega_{2},~ \omega_{3},~ \omega_{4},~ \omega_{5}] $ were chosen as $[1, ~ 100,~ 100, ~ 1,~ 1, ~1]$ for UAV1 to obtain a short collision-free path and $[100, ~1,~ 100, ~ 1,~ 1, ~1]$ for UAV2 and UAV3 to follow their paths while maintaining the formation and avoiding collisions.

In the simulation, we considered two scenarios of construction sites with different sizes of obstacles, illustrating different levels of complexity, to validate the efficiency of the proposed algorithm. 

\subsection{Construction site with big-size obstacles}
In Scenario 1, we considered a construction site with dimensions of $100 m \times 100 m \times 35 m$.  The drones were required to travel from the start location to the goal to perform a site monitoring task. The  start locations of UAVs were set at: $P_1^{start} \!=\! (15; 18.66; 20 ) $, $P_2^{start} \!=\! (10 ; 10 ; 20 ) $, and $P_3^{start} \!=\! (20 ; 10 ; 20 )$. We confirmed the goal poses at $P_1^{goal} \!=\! (85; 88.66; 20) $, $P_2^{goal} \!=\! (80; 80 ; 20) $, and $P_3^{goal} \!=\! (90 ; 80 ; 20 )$. The formation reference was obtained at the same as the target position, i.e., $P_{1_r} \!=\! P_1^{goal} $, $P_{2_r} \!=\! P_2^{goal} $, and $P_{3_r} \!=\! P_3^{goal}$. In the construction site, there were two threat areas modeled as yellow cylinders located at $(40, 40)$ and $(60, 60)$ with a radius of $9$ m. Parameters of the PSO were set as $c_0 \!=\!  0.999$, and $c_1 \!=\! c_2 \!=\!  1.5$. The PSO run with $2000$ particles for $1500$ iterations. Aside from the start and goal nodes, each path was established by $K \!=\!  10$ waypoints.

\begin{figure}[h!]
	\centering
	\includegraphics[width= 0.95 \linewidth]{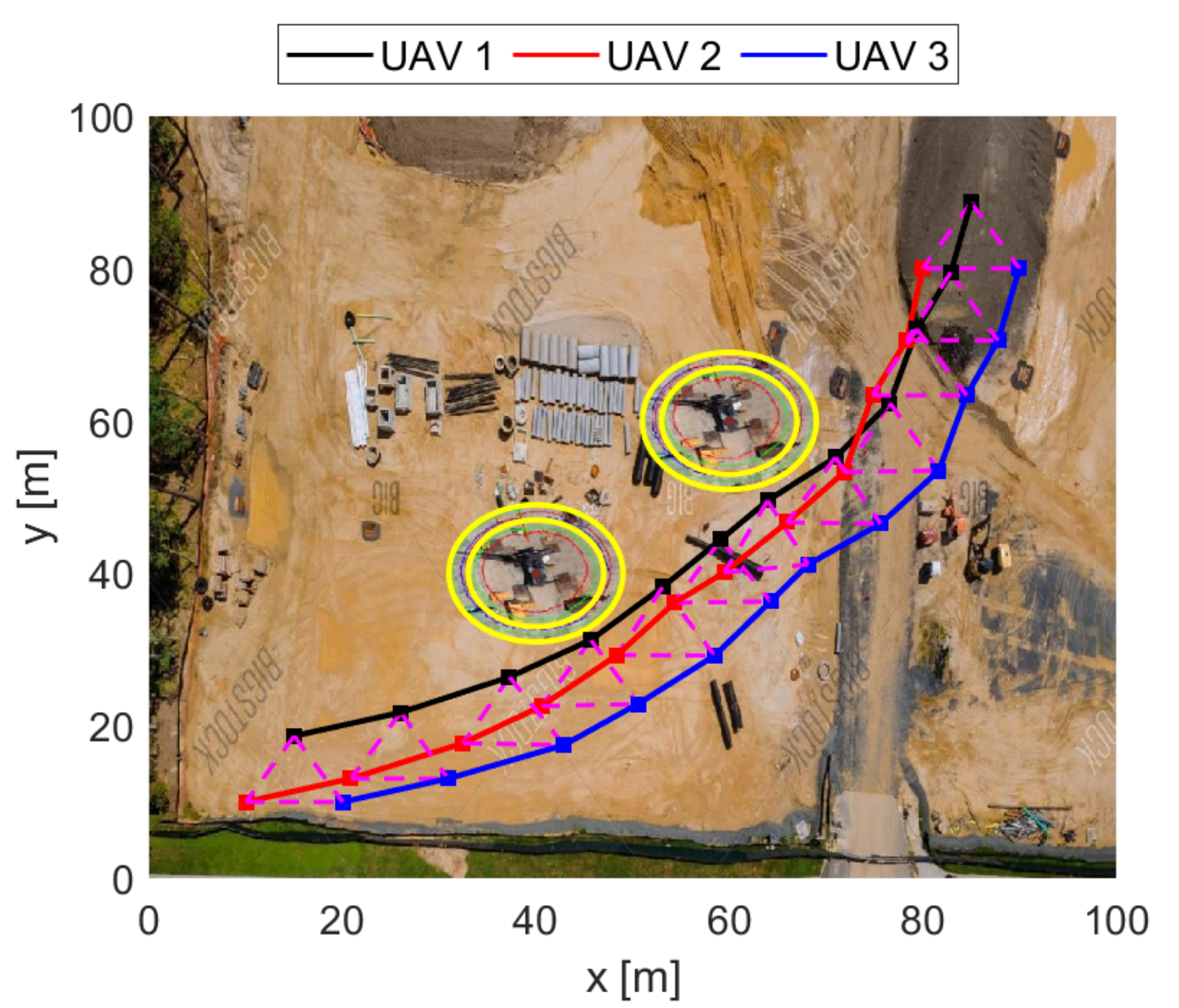}
	\caption{Generated paths for Sencario 1}
	\label{fig4}
\end{figure}

\begin{figure}[h!]
	\centering
	\includegraphics[width= 0.95 \linewidth]{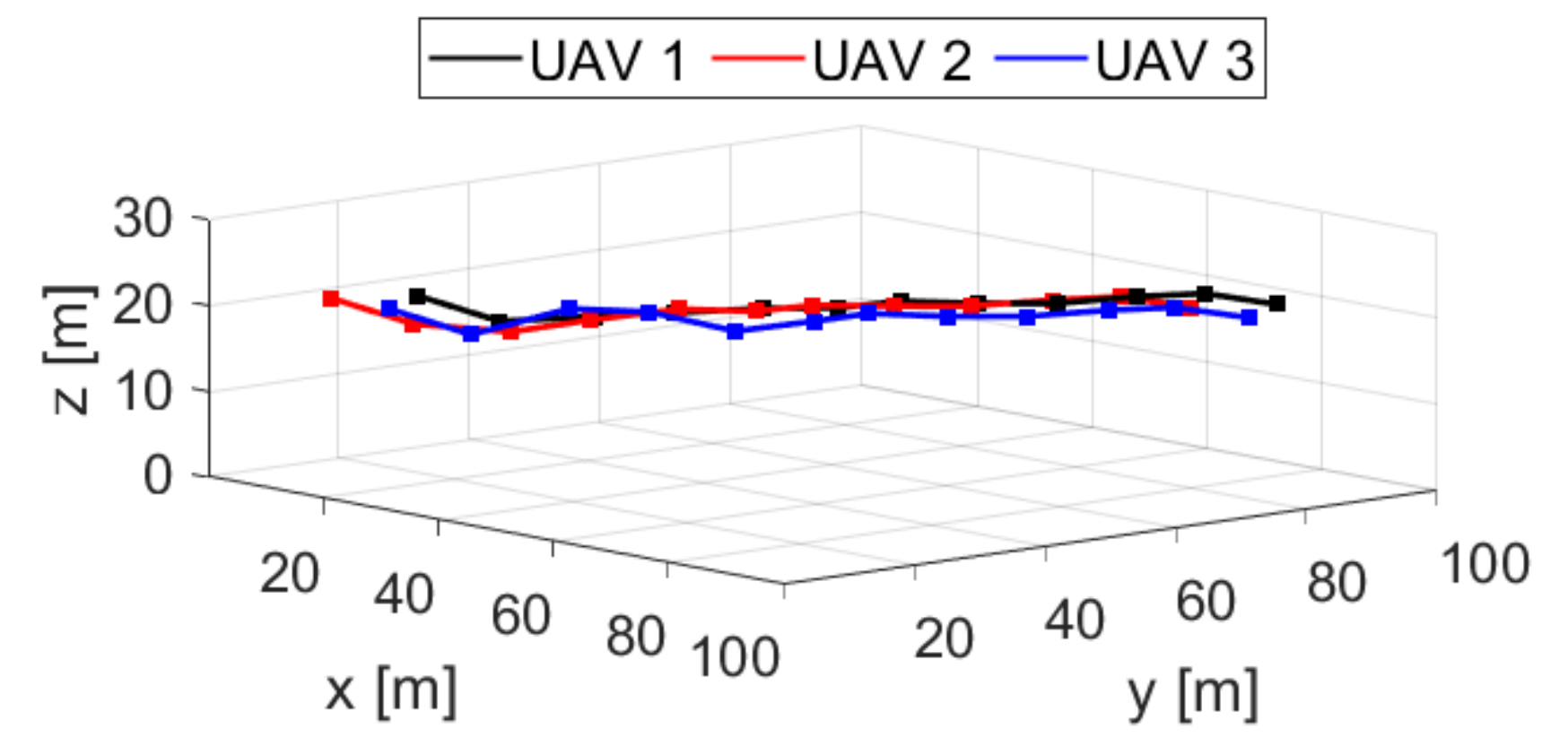}
	\caption{UAVs' altitude}
	\label{fig5}
\end{figure}

The simulation results are shown in Figure \ref{fig4}. The figure shows that collision-free paths are generated. Importantly, an equilateral triangular configuration of UAVs is retained throughout the entire mission, illustrating the efficiency of the proposed approach. The altitude of the UAV team is displayed in Figure \ref{fig5}. The figure demonstrates that the desired height of the vehicles w.r.t. the ground, around $20 ~ m$, is achieved and maintained through the routine, further demonstrating the feasibility of the proposed algorithm.


\vspace{-0.5cm}
\subsection{Construction site with small-size obstacles}
In Scenario 2, we examined the UAV team that has to fly across a construction site to accomplish some tasks. The start and goal locations of the UAVs were chosen to be the same as that of Scenario 1. In Scenario 2, however, two construction cranes as obstacles were located at $(40, 48)$ and $(74, 43)$ with a radius of $4$ m. To enhance the level of complexity, two more virtual obstacles with the same radius were added at $(20, 70)$ and $(70, 60)$. In this simulation, the performance of the proposed stag hunt game-based algorithm was compared with other available techniques that treat the entire UAV fleet as a rigid body, and path planning was achieved for a virtual drone placed at the centre of the formation \cite{chen2015path}. 

\begin{figure}[h!]
	\centering
	\includegraphics[width= 0.95 \linewidth]{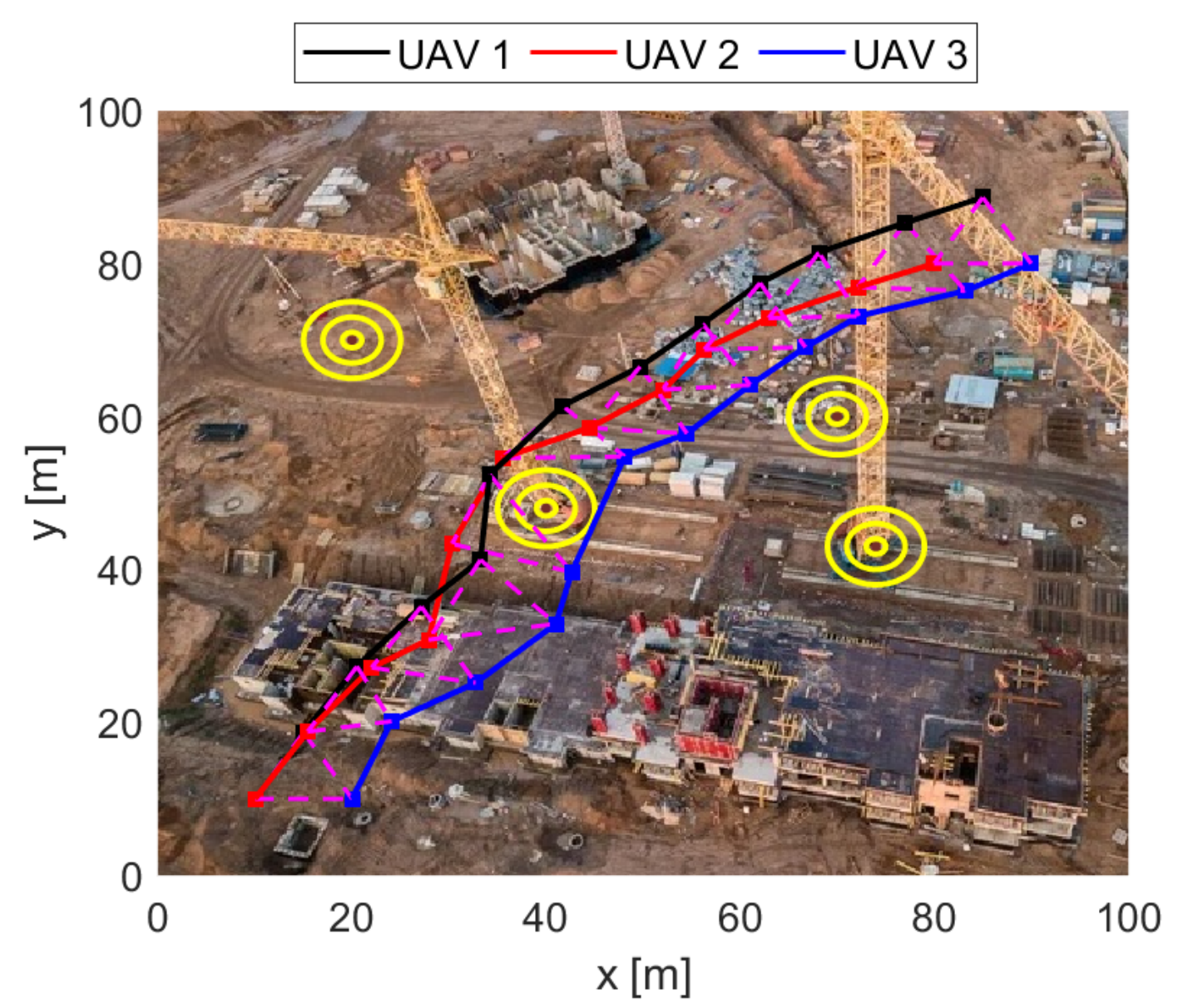}
	\caption{Stag hunt gamed based path}
	\label{fig6}
\end{figure}

\begin{figure}[h!]
	\centering
	\includegraphics[width= 0.95 \linewidth]{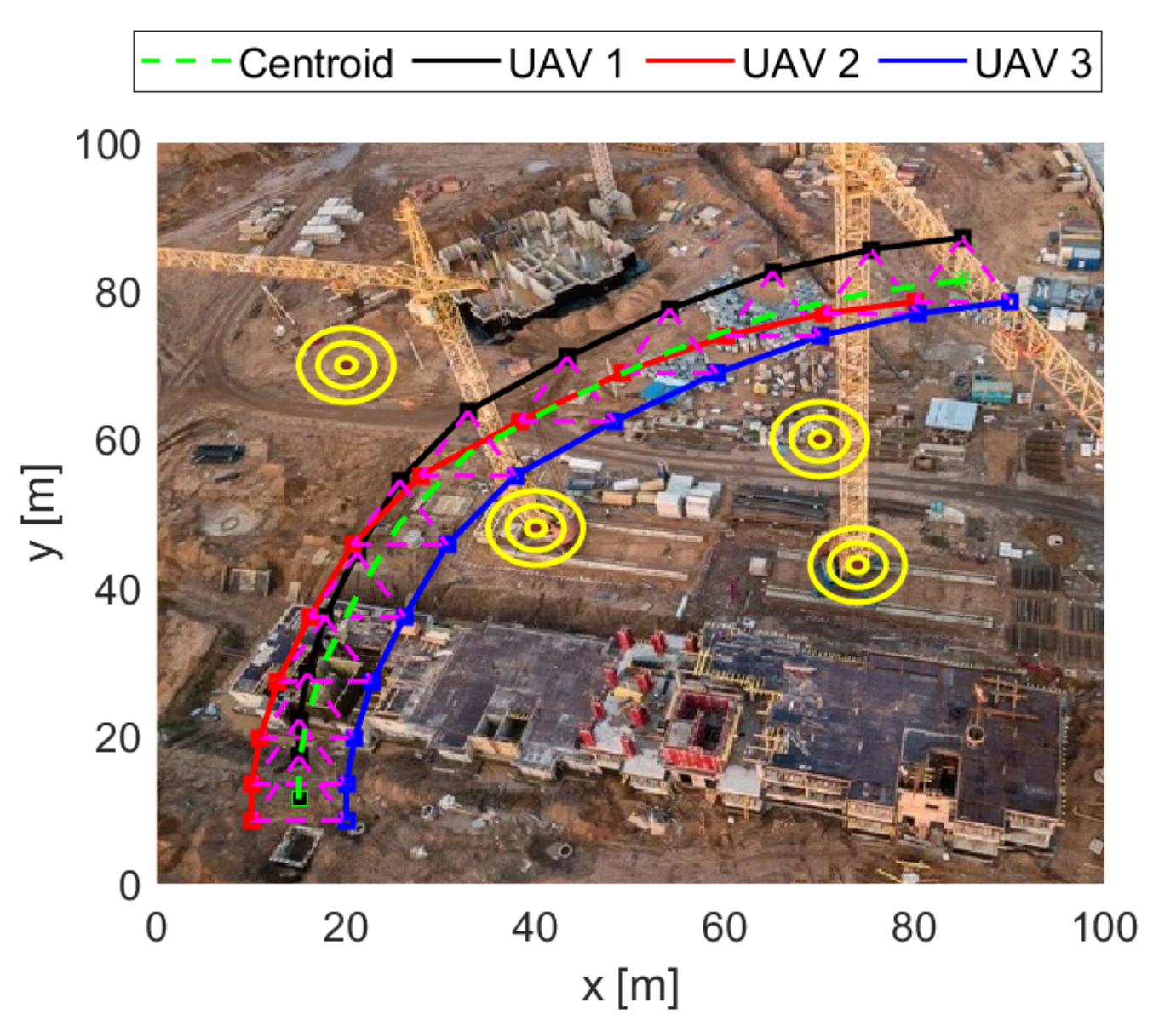}
	\caption{Rigid formation path}
	\label{fig7}
\end{figure}

Figure \ref{fig6} and Figure \ref{fig7} depict the planned stag hunt game-based path and the planned rigid formation path, respectively. It can be seen that both techniques achieve collision-free and formation-preserving routes. However, the whole group of UAVs, using the rigid formation method, travels around obstacles, resulting in long distances. Meanwhile, the proposed game-based approach presents a capability to split and merge the UAV fleet to avoid small-sized threats and reduce the cost. This illustrates the benefit of the proposed algorithm.     

\begin{figure}[h!]
	\centering
	\includegraphics[width= \linewidth]{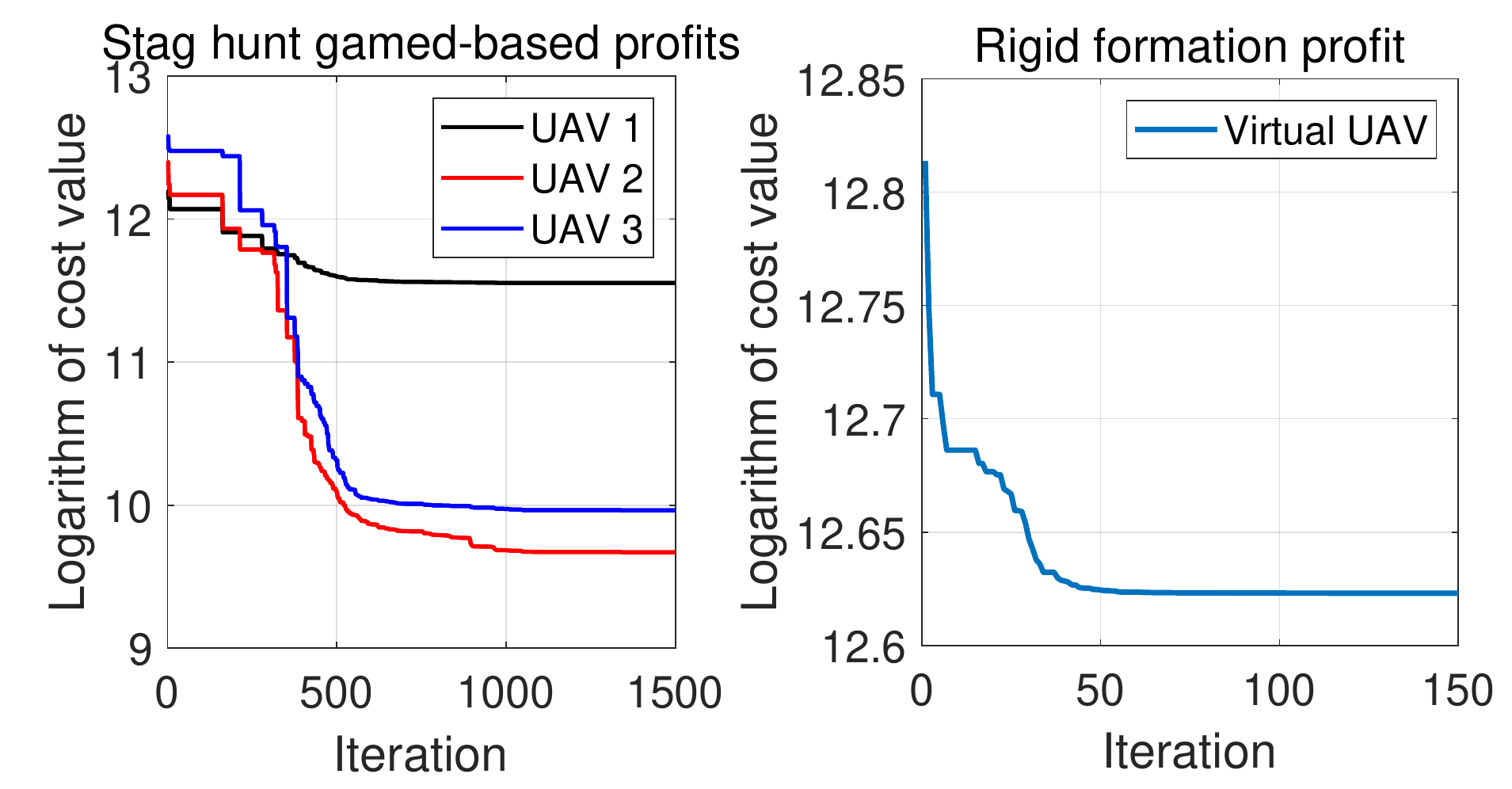}
	\caption{Cost values}
	\label{fig8}
\end{figure}

Figure \ref{fig8} depicts the cost values of all UAVs. As can be seen in the figure, the logarithm of cost values of the UAVs using the stag hunt game-based approach converged to $11.6$, $9.7$, and $10$, after $1000$ iterations, implying that the payoff-dominant equilibrium in the UAV game has been achieved. Compared to the logarithm of the cost value of the virtual UAV using the rigid formation technique, which was about $12.62$, all three UAVs achieved better utility. This confirms that the obtained game-based strategy dominates the rigid formation strategy. Since the proposed method requires a larger number of particles and iterations to achieve the payoff-dominant equilibrium, resulting in a higher computational cost. However, the path planning algorithm is implemented offline, this cost is worth for a better Pareto optimal result.

\vspace{-0.5cm}
\section{Conclusion}
This study has introduced a novel method based on the stag hunt game theory and game-based particle swarm optimization for cooperative UAVs navigating to assemble into a desired formation configuration. The UAV collaborative path planning problem is solved by finding the payoff dominant equilibrium of the game. An optimization framework using PSO was integrated to find the Pareto optimality by minimizing all cost functions simultaneously. Extensive simulation has been conducted to evaluate the performance of the proposed method for various scenarios in monitoring construction sites with multiple UAVs. Our future work will develop receding horizon game theory-based platforms for cooperative UAVs in a dynamic construction environment.

\subsection*{Acknowledgements} This work is supported, in part, by the Vingroup Science and Technology Scholarship (VSTS) Program for Overseas Study for Master’s and Doctoral Degrees.

\bibliography{ISARC}

\end{document}